\documentclass{article}
\usepackage{spconf, graphicx, amsmath, tikz,multirow}
\usetikzlibrary{matrix,chains,positioning,decorations.pathreplacing,arrows}

\usepackage{array,enumitem,hyperref}

\newcolumntype{P}[1]{>{\centering\arraybackslash}m{#1}}
\newcolumntype{L}[1]{>{\arraybackslash}m{#1}}

\makeatletter
\def\blfootnote{\xdef\@thefnmark{}\@footnotetext}
\makeatother

\title{Deep Spectral Convolution Network for HyperSpectral Unmixing}
\name{Savas Ozkan, Gozde Bozdagi Akar}
\address{Middle East Technical University \\ Department of Electrical/Electronics Engineering \\ Ankara, Turkey}
\begin{document}
%
\maketitle
\begin{abstract}

In this paper, we propose a novel hyperspectral unmixing technique based on deep spectral convolution networks (DSCN). Particularly, three important contributions are presented throughout this paper. First, fully-connected linear operation is replaced with spectral convolutions to extract local spectral characteristics from hyperspectral signatures with a deeper network architecture. Second, instead of batch normalization, we propose a spectral normalization layer which improves the selectivity of filters by normalizing their spectral responses. Third, we introduce two fusion configurations that produce ideal abundance maps by using the abstract representations computed from previous layers.  In experiments, we use two real datasets to evaluate the performance of our method with other baseline techniques. The experimental results validate that the proposed method outperforms baselines based on Root Mean Square Error (RMSE).

\end{abstract}

\blfootnote{Email: \url{ozkan.savas@metu.edu.tr}}

\section{Introduction}
\label{sec:intro}

Even though hyperspectral data provides rich content information about the Earth surface and it has been used in a variety of remote sensing applications, the materials exhibited from the surface can be mixed per pixel in different fractions due to the low-spatial resolution of the sensors. Therefore, high dimensional material signatures $E=\{e_1, e_2, ..., e_k\}$ (i.e., endmembers) and their fractions $y=\{y_1, y_2, ..., y_k \}$ (i.e., abundance maps) for each pixel $x$ need to be extracted blindly from data. The mixture of materials can be formulated with a linear model which intuitively defines the necessary parameters for the problem:
\begin{eqnarray}
\label{eqn:bc1}
x = \sum_{k=1}^{K} {e_k \cdot y_k} + \eta, \hspace{2mm}  s.t. \hspace{1.2mm} y_k \geq 0,\hspace{1.2mm}  \sum_{k=1}^{K} {y_k} = 1
\end{eqnarray}

\noindent
where $K$ is the number of materials in the scene and $\eta$ is the random noise to approximate the problem to nonlinearity. Moreover, there are two more constraints which bound the physical properties of the solution.

Primarily, the solutions in literature are highly influenced by the geometrical volume-based assumption where the vertices of data distribution correspond to endmembers, since all data can be reconstructed by the combination of these vertices with different fractions. This derivation is exploited in several approaches as the presence of pure-pixel\cite{heylen2011non, winter1999n}, projection-based~\cite{nascimento2005vertex, harsanyi1994hyperspectral}, kernel-based~\cite{haertel2005spectral, wang2010nonnegative} in literature. However, even if these linear methods work seamlessly to some extent, e.g., on synthetic data / controlled environment, they strive to cope with some cases such as multiple scattering effects, microscopic-level material mixtures and water-absorbed environment on real data~\cite{endnet2017}. Similarly, various solutions based on nonlinear projection~\cite{roweis2000nonlinear, bachmann2006improved} and nonlinear kernel function~\cite{kizel2017stepwise, heylen2015nonlinear} are derived to mitigate these issues.

Very recently, sparse neural networks~\cite{endnet2017, frosti2017} introduce significant performance improvements compared to the traditional blind linear/nonlinear approaches and supervised neural network methods~\cite{plaza2007joint, vcanet2016}. Modifications on the network architecture and loss function constitute the mainstream of these methods. Both methods explain that combination of ReLU activation with batch normalization ultimately improves the sparsity of abundance maps while endmember estimates lead to near-optimum solutions. In addition, ~\cite{endnet2017} introduces a novel loss function which is in accordance with the problem by exploiting spectral angle similarities, regularization layers and additional constrains that boost the sparsity of abundance maps and parameter convergence.

However, as explained~\cite{endnet2017}, these methods should be combined with additional hyperspectral unmixing methods (i.e. even if they outperform the state-of-the-art methods) in order to improve the performance even further. This limitation stems by the fact that high-dimensionality of signatures, shallow network structure and averaging operations in batch normalization. 

\blfootnote{Project Website: \url{https://github.com/savasozkan/dscn}}

In this paper, we propose a deep spectral convolutional network (DSCN) to unmix hyperspectral data with pre-computed endmembers. To address these current limitations, we present several contributions to the architecture as follows:
\begin{itemize}[leftmargin=0.4cm,rightmargin=0cm]
\vspace{-0.2cm}
\item To reduce the adverse effects of high-dimensionality (we will discuss theoretical explanations/observations in detail in Section 2), we replace the fully-connected linear operation with convolutions which enables to extract representative local spectral information from a signature rather than its full version. 

\item Furthermore, use of convolutions allows us to promote a deeper architecture, in other words, a sequence of convolutions which improves the sparsity as well as high-level abstract representation of signatures. 

\item For convolution layers, we replace batch normalization with spectral normalization which aims to improve the spectral selectivity of layers. By this way, more beneficial spectral characteristics can be extracted from hyperspectral data to unmix the fractions of materials.

\item Lastly, two different fusion configurations that estimate ideal abundance map for each pixel are proposed as DSCN-S and DSCN-P which use the representations computed from previous layers. The main difference is that DSCN-S configuration estimates more sparse abundance maps while DSCN-P yields more probabilistic results due to their architecture variations.

\end{itemize}

\section{HyperSpectral Unmixing with Spectral Convolutions}
\label{sec:format}

In this section, we initially formulate the problem to clarify the understandability of the proposed method. Later, we provide theoretical explanations/observations of the modifications that are introduced throughout this paper. Lastly, the details of the method and related information about the architecture are explained.

\subsection{Preliminary} 

\noindent
\textbf{Formulation.} Let $x$ be a pixel of hyperspectral data that is mixed by constituent materials $E$ with various fractions $y$. Generally, two steps should be defined as quantifying of abundance maps $Enc(.)$ and reconstruction of a pixel $Dec(.)$ in order to extract abundance maps $\hat y$ and endmembers $W_d$ respectively from data. Here, $\hat x = Dec(\hat y; W_d)$ is basically equal to vector multiplications as in Eq. 1. On the otherhand, abundance maps $\hat y$ are estimated by $Enc(.)$ as follows:
\begin{eqnarray}
\label{eqn:bc1}
\hat y = Enc(x; \theta_e)  \hspace{2.2mm} s.t. \hspace{1.2mm} \hat y_k \geq 0,\hspace{1.2mm}  \sum_{k=1}^{K} {\hat y_k} = 1
\end{eqnarray}

\noindent
where $\theta_e$ is the set of trainable parameters to obtain optimum abundance map estimates. Remark that we singly focus on to improve $Enc(.)$ step throughout this paper by using pre-computed endmembers $W_d$.

\noindent
\textbf{Impact of Spectral Convolution.}
As explained in detail~\cite{endnet2017}, after the elimination of bias terms, fully-connected linear layer is a simple affine transformation which projects data to a more separable space to ease the estimation process. However, as previously discussed for feature hashing/indexing~\cite{hash2011, pq2011}, when the dimensionality of data increases, irregularity of data leads to holes which hardens to realize an unsupervised method for the problem. A straightforward  solution is to use supervised data to learn a more robust projection to fill these holes~\cite{hash2011}. Another solution is to divide data into several overlapping/non-overlapping parts to increase the representation capacity per element~\cite{pq2011}.

In particular, convolution layer shares similar objective as in the second approach (i.e. small/local parts)~\cite{krizhevsky2012imagenet} and it extracts discriminative responses which indicate the local characteristics of data. Note that we use only 1D convolutions per pixel to identify the spectral characteristics of data, not their spatial information. 

Moreover, since respectively less number of trainable parameters are learned in the convolution layers compared to fully linear operations, a deeper architecture can be promoted in our method. Ultimately, it boosts the discriminative power of the representation by extracting a sequence of abstracts from lower-level to higher ones as explained~\cite{krizhevsky2012imagenet, stacked2006}.

\noindent
\textbf{Spectral Normalization.}
Practically, combination of ReLU with batch normalization enables a network to select the sparse outputs of an affine transformation (i.e. the responses of fully linear / convolution) based on batch characteristics of data. However, this is not completely practical to reveal the latent spectral characteristics of a pixel. 

For this purpose, we utilize spectral normalization with ReLU activation after each convolution layer. Intuitively, spectral normalization normalizes the responses of convolution by regarding their responses for spectral values along with the batch characteristics. By this way, each layer (combination of convolution, spectral norm and ReLU) computes the most representative spectral responses about data while preserving batch characteristic of data. Most relevant normalization type in literature can be seen in~\cite{instance2016} used for style transfer.

\begin{table*}[t]
\begin{center}

\caption{RMSE results on Jasper Ridge dataset. Mean and standard deviation are reported. Best results are shown in bold.}

\begin{tabular}{P{0.8cm} || P{1.45cm} P{1.45cm} P{1.45cm}  P{1.45cm}  P{1.45cm} P{1.45cm} P{1.45cm} P{1.45cm} }
     \hline \hline
   \multirow{2}{*}{} &  \multicolumn{7}{c}{Root Mean Square Error (RMSE) ($\times 10^{-2}$)} \\ \cline{2-9}
      &  VCA  & DMaxD &  SCM &  DgS-NMF & EndNet & EndNet-SPU & EndNet-DSCN-S & EndNet-DSCN-P \\ \hline
     \#1   &  17.68$\pm$6.2  &  17.03$\pm$0.0 &  23.87$\pm$0.0 & 11.66$\pm$0.2 & 10.12$\pm$0.6 & 8.24$\pm$0.4   & 6.04$\pm$1.2 & \textbf{5.77$\pm$0.2} \\
     \#2   &  13.45$\pm$1.9  &  21.34$\pm$0.0 & 13.30$\pm$0.0  & 4.13$\pm$0.0 & 11.48$\pm$0.8 & 6.17$\pm$0.3   &  \textbf{3.96$\pm$0.7} & 4.52$\pm$0.3 \\
     \#3   &  38.93$\pm$7.9  &  14.34$\pm$0.0 & 28.47$\pm$0.0  & 11.13$\pm$0.3  & 9.53$\pm$0.3 & 8.98$\pm$0.2  & \textbf{8.93$\pm$1.5} & 13.07$\pm$0.4 \\
     \#4   &  29.13$\pm$4.2  &  11.21$\pm$0.0 & 19.87$\pm$0.0  & \textbf{5.68$\pm$0.1}  & 12.29$\pm$0.4  & 8.55$\pm$0.1 & 9.31$\pm$1.3 & 14.36$\pm$0.4 \\ \hline
     Avg.  &  24.80$\pm$5.1  &  15.98$\pm$0.0 & 23.04$\pm$0.0 & 8.15$\pm$0.2  & 10.85$\pm$0.6  & 7.96$\pm$0.3 & \textbf{7.06$\pm$0.9} & 9.43$\pm$0.3 \\

  \hline \hline

\end{tabular}
\vspace{-0.2cm}
\label{table:jasper}
\end{center}
\end{table*}

\noindent
\textbf{Fusion Layer.} As indicated~\cite{endnet2017, frosti2017}, combination of ReLU and batch normalization yields robustness for abundance estimates to obtain sparse abundance maps and endmembers. However, for finer abundance values, probabilistic/distance-based approaches might be practical for several datasets as in~\cite{heylen2011non, heylen2015nonlinear}.

For this purpose, throughout the paper, we introduce two difference fusion configurations. If we explain the purpose of fusion layer in detail, the aim is to fuse the responses of spectral convolutions from the previous layers to estimate the true abundance maps per-pixel at the final layer. Note that instead of feeding high-dimensional data directly to a fully-connected linear layer as in~\cite{endnet2017, frosti2017}, the proposed method reduces the dimension of data iteratively while enriching the representation capacity of the input with sparse transformations (i.e. convolutions) for both configurations. 

First configuration, i.e. DSCN-S, computes the ideal abundance values for each pixel with the combination of fully-connected linear, batch norm, ReLU and l1-norm layers by taking the hidden representation computed from the spectral convolutions as an input. Note that due to joint usage of ReLU and batch normalization, it is expected that the estimated abundance maps are quite sparse. 

Second configuration, i.e. DSCN-P, consists of fully-connected linear and softmax activation layers. The softmax activation function response is as follows: 
\begin{eqnarray}
\label{eqn:bc5}
\begin{aligned}
softmax(h) =    \frac{e^{h}}{\sum_{k=1}^{K} e^{h_k}}
\end{aligned}
\end{eqnarray}

\noindent
where $h$ is the outputs of fully-connected linear layer. Due to the architecture, this configuration yields probabilistic results and finer abundance maps for highly-mixtured scenes. Lastly, both of these configuration layers is the final layer of $Enc(.)$.

\begin{figure}
\centering
\includegraphics[scale=0.5]{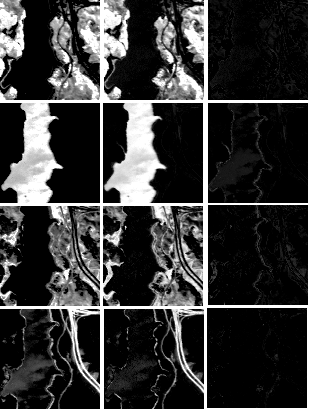}
\vspace{-0.2cm}
\caption{Visualization of the results of the proposed method on Jasper Ridge dataset for each material. From left to right, ground truth, estimated abundance map, absolute difference respectively.}
\vspace{-0.2cm}
\label{fig:jasper}
\end{figure}

\subsection{Deep Spectral Convolution Network}

\noindent
\textbf{Architecture.}
First, a pixel is filtered by two consecutive spectral convolution blocks. Note that the number of blocks and inner structure can still be tuned for different datasets (i.e. deeper networks). Each block consists of spectral normalization and ReLU activation layers after a spectral convolution. To reduce the dimensionality of responses, maxpool layer is exploited at each block. To this end, these blocks ultimately behave like a feature extractor. 

At the third block, batch normalization and ReLU are utilized with spectral convolution to determine the responses based on only their batch characteristics. This implicitly corresponds to the mutual distribution of data as in~\cite{heylen2011non, heylen2015nonlinear}. This is critical since the final convolution block reduces the depth size regarding to the overall data batch characteristic. Lastly, fusion layer (i.e. either DSCN-S or DSCN-P) is used to compute abundance maps for a pixel. 

\noindent
\textbf{Learning.} For parameter optimization, we use the loss function that is proposed in~\cite{endnet2017} and the parameters $\theta_e$ are updated by back-propagation scheme. This full loss function is written as:
\begin{eqnarray}
\label{eqn:bc5}
\begin{aligned}
\mathcal{L} =    - \lambda_1 \text{D}_{\text{KL}}\big( \hspace{0.1mm} 1.0 || C(x, \hat{x}) \hspace{0.1mm}  \big) + \lambda_2 \|  \hat y \|_1 +  \lambda_3 \| \theta_e \|_2
\end{aligned}
\end{eqnarray}

\noindent
where $\lambda_1$, $\lambda_2$ and $\lambda_3$ are set to 10, 0.4 and $10^{-5}$ respectively. $\text{D}_{\text{KL}}(.)$ is the Kullback-Leibler divergence term and $C(., .)$ is the normalized SAD score between the original and reconstructed version of signatures~\cite{endnet2017}.

Note that finetuning of pre-computed endmember $W_d$ is not allowed during the training. Moreover, unlike~\cite{endnet2017}, denoising autoencoder scheme is not used for the method, since our aim is to obtain actual/finer abundance values rather than coarse estimation of constituent endmembers from data. 

Adam stochastic optimizer~\cite{kingma2014adam} is used with the previously explained parameter settings~\cite{endnet2017}. The number of iteration is set to 5K and the parameters are randomly initialized. Codes are implemented on Python by extensively leveraging Tensorflow framework.

\section{Experiments}
\label{sec:pagestyle}

\subsection{Datasets, Evaluation Metric and Baselines}

\begin{table*}[t]
\begin{center}

\caption{RMSE results on Urban dataset. Mean and standard deviation are reported. Best results are shown in bold.}

\begin{tabular}{P{0.8cm} || P{1.45cm} P{1.45cm} P{1.45cm}  P{1.45cm}  P{1.45cm} P{1.45cm} P{1.45cm} P{1.45cm} }
     \hline \hline
   \multirow{2}{*}{} &  \multicolumn{7}{c}{Root Mean Square Error (RMSE) ($\times 10^{-2}$)} \\ \cline{2-9}
      &  VCA  & DMaxD &  SCM &  DgS-NMF & EndNet & EndNet-SPU & EndNet-DSCN-S & EndNet-DSCN-P \\ \hline
     \#1   &  42.14$\pm$7.2  &  30.68$\pm$0.0 &  32.79$\pm$0.0 & 13.18$\pm$0.1 & 13.04$\pm$0.3 & 10.41$\pm$0.2   & 12.85$\pm$1.3 & \textbf{9.87$\pm$0.2} \\
     \#2   &  48.46$\pm$5.6  &  47.26$\pm$0.0 & 36.25$\pm$0.0  & 12.95$\pm$0.0 & 14.43$\pm$0.3 & 12.24$\pm$0.3   & 13.67$\pm$1.5 & \textbf{12.08$\pm$0.3} \\
     \#3   &  17.18$\pm$3.7  &  26.71$\pm$0.0 & 32.61$\pm$0.0  & 9.57$\pm$0.1  & 8.71$\pm$0.5 & 8.35$\pm$0.3  & 8.53$\pm$0.7 & \textbf{7.54$\pm$0.1} \\
     \#4   &  16.94$\pm$2.1  &  19.49$\pm$0.0 & 32.86$\pm$0.0  & 6.27$\pm$0.0  & 7.59$\pm$0.2  & \textbf{5.92$\pm$0.1} & 7.03$\pm$0.9 & 6.65$\pm$0.1 \\ \hline
     Avg.  &  31.18$\pm$4.7  &  31.04$\pm$0.0 & 33.59$\pm$0.0 & 10.49$\pm$0.1  & 10.94$\pm$0.4  & 9.23$\pm$0.2 & 10.52$\pm$1.1 & \textbf{9.04$\pm$0.2} \\

  \hline \hline

\end{tabular}
\vspace{-0.3cm}
\label{table:urban}
\end{center}
\end{table*}

To make fair and realistic comparisons, we evaluate the proposed method on two real datasets, namely \textit{Jasper Ridge}~\cite{zhu2014spectral} and \textit{Urban}~\cite{zhu2014spectral}, which are extensively used in literature. Briefly, for \textit{Jasper Ridge} dataset, the spectral and spatial resolutions are 198 and 100 $\times$ 100, respectively. There are four main materials: Tree (\#1), Water (\#2), Soil (\#3) and Road (\#4). The spatial resolution of \textit{Urban} dataset is 307 $\times$ 307 and its spectral resolution is 162. Similarly, it has four constituent materials in the scene: Asphalt (\#1), Grass (\#2), Tree (\#3) and Roof (\#4). For reliable assessments, tests are repeated 20 times for each method, thus mean and standard deviation of the results are reported.

Furthermore, we compare the performance of the method with several baseline algorithms such as VCA~\cite{nascimento2005vertex}, DMaxD~\cite{heylen2011non}, SCM~\cite{zhou2016spatial}, DgS-NMF~\cite{zhu2014spectral}, EndNet~\cite{endnet2017} and EndNet-SPU~\cite{endnet2017}. For a performance metric, we utilize Root Mean Square Error (RMSE) to measure the error between estimated abundance maps and ground truth. To preserve non-linearity for VCA and DMaxD abundance estimates, we compute their endmember estimates with Multilinear Mixing Model (MLM)~\cite{heylen2015nonlinear} throughout the experiments. Lastly, the proposed unmixing method (i.e. either DSCN-S or DSCN-P) aims to improve the abundance map results of endmembers estimated by EndNet~\cite{endnet2017} which recently achieves the state-of-the-art performance in literature. You can find further detail about EndNet from~\cite{endnet2017}.

\subsection{Experimental Results}

\noindent
\textbf{Jasper Ridge.} Experimental results for this dataset are illustrated in Table~\ref{table:jasper}. From these results, the sparse version of the proposed method (i.e. DSCN-S) achieves the best overall accuracy. It approximately introduces 1\% improvements to the second best result which is obtained by EndNet-SPU combination. As identified~\cite{endnet2017},  Soil (\#3) and Road (\#4) materials are highly correlated and it is only practical to quantify the fractions with supervised data or spatial reasoning as in DgS-NMF method. For Tree (\#1) and Water (\#2) materials in particular, the proposed method nearly attains ideal abundance performance for the materials. 

In addition, Fig.~\ref{fig:jasper} shows the quantitative results of the method (DSCN-S) for each materials (i.e. each row). Perceptually impressive results are obtained especially for Soil and Water. Similarly, the error concentrates  at the boundaries of water-ground as well as road-soil intersections.

However, there is an issue for DSCN-S that the variances in the accuracies are a bit high while DSCN-P obtains more stable results. The main drawback arises primarily due to the lack of parameter convergence to a global solution for every initialization. We believe that this issue can be reduced by the detail experiments on the network architecture.

\noindent
\textbf{Urban.} Table~\ref{table:urban} shows the experimental results on Urban dataset. The probabilistic configuration of the proposed method, DSCN-P, obtains the best overall results with small improvements over EndNet-SPU. 

Note that the actual abundance maps of data is quite dense (i.e. highly-mixtured), thus the probabilistic version can be more appropriate for this case. The experiment results also support this assumption by yielding better performance. Lastly, the variation in the scores is still an issue for DSCN-S while DSCN-P generates more consistent results.

\section{Conclusion}
\label{sec:typestyle}

In this paper, we propose a deep spectral convolution network to unmix hyperspectral data with pre-computed endmembers. Throughout the paper, we introduce three critical contributions for the unmixing problem. First, instead of a single layer fully-connected linear operation, a network that is composed of several spectral convolution layers with a deeper architecture is proposed. Later, we present a novel spectral normalization layer that is able to normalize responses of filters to improve the selectivity of layers. Lastly, we introduce two configurations for the fusion of the responses of previous layers and the computation of abundance maps. The experimental results validate that the proposed method in this paper obtains the new state-of-the-art performance on two real datasets.

\section{ACKNOWLEDGMENTS}
\label{sec:copyright}

We gratefully acknowledge the support of NVIDIA Corporation with the donation of Quadro P5000 used for this research.

\bibliographystyle{IEEEtrans}
\bibliography{Template}

\end{document}